\pdfoutput=1

\documentclass[11pt]{article}
\usepackage[]{acl2021}
\usepackage{mathtools}
\usepackage{makecell}
\usepackage{threeparttable}
\usepackage{bm}

\usepackage{setspace}
\usepackage{enumitem}

\usepackage{times}
\usepackage{url}
\usepackage{latexsym}
\usepackage{multicol}
\usepackage{multirow}
\usepackage{booktabs}

\usepackage{todonotes}
\usepackage{comment}

\title{An Empirical Survey of Data Augmentation \\for Limited Data Learning in NLP}
\author{
 Jiaao Chen$^\diamond$\thanks{\ \ Equal contribution.} \quad Derek Tam$^{\dagger*}$ \quad Colin Raffel$^{\dagger}$ \quad Mohit Bansal$^{\dagger}$ \quad Diyi Yang$^\diamond$ \\
 $^\diamond$Georgia Institute of Technology \quad $^\dagger$UNC Chapel Hill \\
\texttt{\{jchen896,dyang888\}@gatech.edu} \\ 
\texttt{ \{dtredsox, craffel, mbansal\}@cs.unc.edu}
}

\begin{document}
\maketitle
\begin{abstract}

NLP has achieved great progress in the past decade through the use of neural models and large labeled datasets.
The dependence on abundant data prevents NLP models from being applied to low-resource settings or novel tasks where significant time, money, or expertise is required to label massive amounts of textual data. Recently, data augmentation  methods have been explored as a means of improving data efficiency in NLP.
To date, there has been no systematic empirical overview of data augmentation for NLP in the limited labeled data setting, making it difficult to understand which methods work in which settings.
In this paper, we provide an empirical survey of recent progress on data augmentation for NLP in the limited labeled data setting, summarizing the landscape of methods (including token-level augmentations, sentence-level augmentations, adversarial augmentations and hidden-space augmentations) and carrying out experiments on 11 datasets covering topics/news classification, inference tasks, paraphrasing tasks, and single-sentence tasks. Based on the results, we draw several conclusions to help practitioners choose appropriate augmentations in different settings and discuss the current challenges and future directions for limited data learning in NLP.
\end{abstract}

\section{Introduction}

Deep learning methods have achieved strong performance on a wide range of supervised learning tasks \cite{Sutskever2014Sequence,deng2013recent,minaee2021deep}.
Traditionally, these results were attained through the use of large, well-labeled datasets.
This make them challenging to apply in settings where collecting a large amount of high-quality labeled data for training is expensive. Moreover, given the fast-changing nature of real-world applications, it is infeasible to relabel every example whenever new data comes in.
This highlights a need for learning algorithms that can be trained with a limited amount of labeled data.

There has been a substantial amount of research towards learning with limited labeled data for various tasks in the NLP community.
One common approach for mitigating the need for labeled data is \textbf{data augmentation}.
Data augmentation \cite{feng2020survey} generates new data by modifying existing data points through transformations that are designed based on prior knowledge about the problem's structure \cite{Yang2015That,wei2019eda}.
This augmented data can be generated from labeled data, and then directly used in supervised learning \cite{wei2019eda}, or in semi-supervised learning for unlabeled data through consistency regularization \cite{xie2019unsupervised} (``consistency training''). While various approaches have been proposed to tackle learning with limited labeled data --- including unsupervised pre-training \cite{peters2018deep,devlin2018bert,raffel2019exploring}, multi-task learning \cite{glorot2011domain,liu2017adversarial,augenstein2018multitask}, semi-supervised learning \cite{zhu2005semi,Chapelle2009Semi,miyato2016adversarial,xie2019unsupervised}, and few-shot learning \cite{deng2019low} --- in this work, we focus on and compare different data augmentation methods and their application to supervised and semi-supervised learning.

In this survey, we comprehensively review and perform experiments on recent data augmentation techniques developed for various NLP tasks. Our contributions are three-fold: 
(1) summarize and categorize recent methods in textual data augmentation; (2) compare different data augmentation methods through experiments with limited labeled data in supervised and semi-supervised settings on 11 NLP tasks, and (3) discuss current challenges and future directions of data augmentation, as well as learning with limited data in NLP more broadly.
Our experimental results allow us to conclude that no single augmentation works best for every task, but (i) token-level augmentations work well for supervised learning, (ii) sentence-level augmentation usually works the best for semi-supervised learning, and (iii) augmentation methods can sometimes hurt performance, even in the semi-supervised setting.

\paragraph{Related Surveys.}
Recently, several surveys also explore the data augmentation techniques for NLP \cite{journals/corr/abs-2010-12309,feng2020survey}.  \citet{journals/corr/abs-2010-12309} provide a broad overview of techniques for NLP in low resource scenarios and briefly cover data augmentation as one of several techniques. In contrast, we focus on data augmentation and provide a more comprehensive review on recent data augmentation methods in this work. While \citet{feng2020survey} also survey task-specific data augmentation approaches for NLP, our work summarizes recent data augmentation methods in a more fine-grained categorization. We also focus on their application to learning from limited data by providing an \textbf{empirical study} over different augmentation methods on various benchmark datasets in both supervised and semi-supervised settings, so as to hint data augmentation selections in future research.

\section{Data Augmentation for NLP}
\label{sec:DA} 
Data augmentation increases both the amount (the number of data points) and the diversity (the variety of data) of a given dataset \cite{cubuk2019autoaugment}. Limited labeled data often leads to overfitting on the training set 
and data augmentation works to alleviate this issue by manipulating data either automatically or manually to create additional augmented data.Such techniques have been widely explored in the computer vision field, with methods like geometric/color space transformations~\cite{Simard2003BestPF,Krizhevsky2012ImageNet,taylor2017improving}, mixup~\cite{zhang2017mixup}, and random erasing~\cite{zhong2017random,devries2017improved}. 
Although the discrete nature of textual data and its complex syntactic and semantic structures make finding label-preserving transformation more difficult, there nevertheless exists a wide range of methods for augmenting text data that in practice preserve labels. 
In the following subsections, we describe four broad classes of data augmentation methods:

\begin{table*}[ht!]
\centering
\small
\begin{threeparttable}
\begin{tabular}{rcccl} 
\toprule
\textbf{Methods}                                                                              & \textbf{Level}                                                       &  \textbf{Diversity} & \textbf{Tasks}                                                                                                                        & \textbf{Related Work} \\ \midrule  \midrule 
\begin{tabular}[c]{@{}c@{}}Synonym \\replacement \end{tabular}                                                               & Token                                                           & Low       & \begin{tabular}[c]{@{}l@{}}Text classification\\ Sequence labeling \end{tabular}                                              & \small\begin{tabular}[c]{@{}l@{}} \citet{kolomiyets-etal-2011-model}, \citet{Zhang2015Character},\\  \citet{Yang2015That}, \citet{miao2020snippext},\\ \citet{wei2019eda}  
\end{tabular}       \\ \midrule  
\begin{tabular}[c]{@{}r@{}}Word  replacement\\  via LM\end{tabular} & Token                                                         & Medium    & \begin{tabular}[c]{@{}l@{}}Text classification\\ Sequence labeling\\ Machine translation\end{tabular}                        & \small\begin{tabular}[c]{@{}l@{}} \citet{kolomiyets-etal-2011-model},  \citet{zhu2019soft}\\ \citet{kobayashi2018contextual}, \citet{wu2019conditional}\\\citet{fadaee-etal-2017-data}  \end{tabular}      \\ \midrule  
\begin{tabular}[c]{@{}r@{}}Random insertion,\\deletion, swapping\end{tabular}  & Token                                                          & Low       & \begin{tabular}[c]{@{}l@{}}Text classification\\ Sequence labeling\\ Machine translation \\ Dialogue generation \end{tabular}                        & \small\begin{tabular}[c]{@{}l@{}} \citet{iyyer-etal-2015-deep},  \citet{DBLP:journals/corr/XieWLLNJN17}\\ \citet{artetxe2017unsupervised}, \citet{lample2018unsupervised}\\  \citet{xie2019unsupervised}, \citet{wei2019eda} \end{tabular}     \\ \midrule  
\begin{tabular}[c]{@{}r@{}}Compositional \\Augmentation\end{tabular}  & Token                                                            & High       & \begin{tabular}[c]{@{}l@{}}Semantic Parsing\\ Sequence labeling\\ Language modeling \\ Text generation \end{tabular}                        & \small\begin{tabular}[c]{@{}l@{}} \citet{jia-liang-2016-data} , \citet{andreas-2020-good}\\  \citet{nye2020learning}, \citet{feng2020genaug}\\ \citet{furrer2020compositional} , \citet{guo-etal-2020-sequence} \end{tabular}     \\ \midrule  \midrule 
Paraphrasing                                                                         & Sentence                                                        & High      & \begin{tabular}[c]{@{}l@{}}Text classification\\ Machine translation\\ Question answering\\ Dialogue generation \\Text summarization \end{tabular} & \small\begin{tabular}[c]{@{}l@{}} \citet{DBLP:journals/corr/abs-1804-09541}, \citet{xie2019unsupervised}\\ \citet{Chen2020ControllablePA}, \citet{he2019revisiting} \\\citet{chen2020mixtext}, \citet{cai-etal-2020-data} \end{tabular}         \\ \midrule  
\begin{tabular}[c]{@{}r@{}} Conditional\\ generation  \end{tabular}                                                             & Sentence                                                         & High      & \begin{tabular}[c]{@{}l@{}}Text classification\\ Question answering\end{tabular}                                             & \small\begin{tabular}[c]{@{}l@{}} \citet{anabytavor2019data}, \citet{kumar2020data}\\  \citet{zhang2019addressing},   \citet{yang2020gdaug}\end{tabular}         \\ \midrule  \midrule  
\begin{tabular}[c]{@{}r@{}} White-box\\ attack  \end{tabular}                                                                  & \begin{tabular}[c]{@{}l@{}}Token or\\ Sentence\end{tabular}                                                         & Medium    & \begin{tabular}[c]{@{}l@{}}Text classification\\ Sequence labeling\\ Machine translation\end{tabular}                                              & \small\begin{tabular}[c]{@{}l@{}} \citet{miyato2016adversarial}, \citet{ebrahimi2017hotflip}\\  \citet{ebrahimi2018adversarial}, \citet{Cheng_2019}, \\ \citet{chen-etal-2020-seqvat}\end{tabular}         \\ \midrule  
\begin{tabular}[c]{@{}r@{}} Black-box\\ attack \end{tabular}                                                                   & \begin{tabular}[c]{@{}l@{}}Token or\\ Sentence\end{tabular}    & Medium    & \begin{tabular}[c]{@{}l@{}}Text classification \\Sequence labeling \\ Machine translation\\ Textual entailment \\Dialogue generation \\ Text Summarization \end{tabular}                                 & \small\begin{tabular}[c]{@{}l@{}}\citet{jia2017adversarial}\\  \citet{belinkov2017synthetic}, \citet{zhao2017generating}\\  \citet{ribeiro-etal-2018-semantically}, \citet{mccoy2019right}\\ \citet{min2020syntactic}, \citet{tan-etal-2020-morphin} \end{tabular}     \\ \midrule  \midrule 
\begin{tabular}[c]{@{}r@{}} Hidden-space\\ perturbation \end{tabular}                                                                 & \begin{tabular}[c]{@{}l@{}}Token or\\ Sentence\end{tabular}                                                           & High      & \begin{tabular}[c]{@{}l@{}}Text classification\\ Sequence labeling \\ Speech recognition \end{tabular}                     &
\small\begin{tabular}[c]{@{}l@{}} \citet{Hsu_2017}, \citet{Hsu_2018}\\ \citet{Wu2019}, \citet{chen2021hiddencut}\\  \citet{malandrakis-etal-2019-controlled}, \citet{shen2020simple}\end{tabular}\\ \midrule 
\begin{tabular}[c]{@{}r@{}} Interpolation \end{tabular}                                                                 & Token                                                          & High      & \begin{tabular}[c]{@{}l@{}}Text classification\\ Sequence labeling \\ Machine translation\end{tabular}                     &
\small\begin{tabular}[c]{@{}l@{}}\citet{miao2020snippext}, \citet{chen2020mixtext}\\  \citet{cheng-etal-2020-advaug}, \citet{chen-etal-2020-local} \\ \citet{guo-etal-2020-sequence} \end{tabular}\\ \bottomrule 
\end{tabular}
\caption{\label{da-sum} \small Overview of different data augmentation techniques in NLP. 
Diversity refers to the difference of augmented data from existing data and the amount of different augmented data could be generated.}
\end{threeparttable}
\end{table*}

\subsection{Token-Level Augmentation}
\label{subsec:reg-DA}

Token-level augmentations manipulate words and phrases in a sentence to generate augmented text while ideally retaining the semantic meaning and labels of the original text. 

\paragraph{Designed Replacement.}
Intuitively, the semantic meaning of a sentence remains unchanged if some of its tokens are replaced with other tokens that have the same meaning. A simple approach is to fetch synonyms as words for substitutions \cite{kolomiyets-etal-2011-model,Yang2015That,Zhang2015Character,wei2019eda,miao2020snippext}. The synonyms are discovered based on pre-defined dictionaries such as WordNet \cite{kolomiyets-etal-2011-model}, or similarities in word embedding space \cite{Yang2015That}. However, improvements from this technique are usually minimal \cite{kolomiyets-etal-2011-model} and in some cases, performance may even degrade \cite{Zhang2015Character}. A major drawback stems from the lack of contextual information when fetching synonyms---especially for words with multiple meanings and few synonyms. To resolve this, language models (LMs) have been used to replace the sampled words given their context \cite{kolomiyets-etal-2011-model,fadaee-etal-2017-data,kobayashi2018contextual,kumar2020data}. Other work preserves the labels of the text by 
conditioning on the label when generating the LMs' predictions \cite{kobayashi2018contextual,wu2019conditional}.
In addition, different sampling strategies for word replacement have been explored.
For example, instead of sampling one specific word from candidates by LMs, \citet{zhu2019soft} propose to compute a weighted average over embeddings of possible words predicted by LMs as the replaced input since the averaged representations could augment text with richer information.

\paragraph{Random Insertion, Replacement, Deletion and Swapping.}
While well-designed local modifications can preserve the syntax and semantic meaning of a sentence \cite{niu2018adversarial}, random local modifications such as deleting certain tokens \cite{iyyer-etal-2015-deep,wei2019eda,miao2020snippext}, inserting random tokens \cite{wei2019eda,miao2020snippext}, replacing non-important tokens with random tokens \cite{DBLP:journals/corr/XieWLLNJN17,xie2019unsupervised,niu2018adversarial} or randomly swapping tokens in one sentence \cite{artetxe2017unsupervised,lample2018unsupervised,wei2019eda,miao2020snippext} can preserve the meaning in practice. Different kinds of operations can be further combined \cite{wei2019eda}, where each example is randomly augmented with one of insertion, deletion, and swapping. 
These noise-injection methods can efficiently be applied to training, and show improvements when they augment simple models trained on small training sets. However, the improvements might be unstable due to the possibility that random perturbations change the meanings of sentences \cite{niu2018adversarial}. Also, finetuning large pre-trained models on specific tasks might attenuate improvements due to preexisting generalization abilities of the model \cite{shleifer2019low}. 

\noindent
\textbf{Compositional Augmentation.} To increase the compositional generalization abilities of models, recent efforts have also focused on compositional augmentations \cite{jia-liang-2016-data,andreas-2020-good} where different fragments from different sentences are re-combined to create augmented examples. Compared to random swapping, compositional augmentation often requires more carefully-designed rules such as lexical overlap \cite{andreas-2020-good}, neural-symbolic stack machines \cite{chen2020compositional}, and neural program
synthesis \cite{nye2020learning}. With the potential to greatly improve the generalization abilities to out-of-distribution data, compositional augmentation has been utilized in sequence labeling \cite{guo-etal-2020-sequence}, semantic parsing \cite{andreas-2020-good,nye2020learning,furrer2020compositional}, language modeling \cite{andreas-2020-good,shaw2020compositional}, and text generation \cite{feng2020genaug}.

\subsection{Sentence-Level Augmentation}
\label{subsec:sent-DA}

Instead of modifying tokens, sentence-level augmentation modifies the entire sentence at once.

\paragraph{Paraphrasing.} Paraphrasing has been widely adopted as a data augmentation technique in various NLP tasks \cite{DBLP:journals/corr/abs-1804-09541,xie2019unsupervised,kumar-etal-2019-submodular,he2019revisiting,chen2020semi,chen2020mixtext,cai-etal-2020-data}, as it generally provides more diverse augmented text with different word choices and sentence structures while preserving the meaning of the original text. 
The most popular is round-trip translation \cite{Sennrich2015Improving,edunov2018understanding}, a pipeline which first translates sentences into certain intermediate languages and then translates them back to generate paraphrases.
Translating through intermediate languages with different vocabulary and linguistic structures can generate useful paraphrases.
To ensure the diversity of augmented data, sampling and noisy beam search can also be adopted during the decoding stage \cite{edunov2018understanding}. Other  work focuses on directly training end-to-end models to generate paraphrases \cite{prakash2016neural}, and further augments the decoding phase with syntactic information \cite{iyyer2018adversarial,Chen2020ControllablePA}, latent variables~\cite{gupta2017deep}, and sub-modular objectives \cite{kumar-etal-2019-submodular}. 

\paragraph{Conditional Generation.}
Conditional generation methods generate additional text from a language model, conditioned on the label. After training the model to generate the original text given the label, the model can generate new text \cite{anabytavor2019data,zhang2019addressing,kumar2020data,yang2020gdaug}.  An extra filtering process is often used to ensure high-quality augmented data. For example, in text classification, \citet{anabytavor2019data} first fine-tune  GPT-2 \cite{radford2019language} with the original examples prepended with their labels, and then generate augmented examples by feeding the fine-tuned model certain labels. Only confident examples as judged by a baseline classifier trained on the original data are kept. Similarly, new answers are generated on the basis of given questions in question answering and are filtered by customized metrics like question answering probability \cite{zhang2019addressing} and n-gram diversity \cite{yang2020gdaug}. Generative models used in this setting have been based on conditional VAE \cite{bowman2015generating,hu2017controlled,guu2017generating,malandrakis-etal-2019-controlled}, GAN \cite{iyyer2018adversarial,xu2018dpgan} or pre-trained language models like GPT-2 \cite{anabytavor2019data,kumar2020data}. Overall, these conditional generation methods can create novel and diverse data that might be unseen in the original dataset, but require significant training effort.

\subsection{Adversarial Data Augmentation}
\label{subsec:adv-DA}

Adversarial methods create augmented examples by adding adversarial perturbations to the original data, which dramatically influences the model's predictions and confidence without changing human judgements. These adversarial examples \cite{morris-etal-2020-textattack-framework,zeng2020openattack} could be leveraged in adversarial training \cite{goodfellow2014explaining} to increase neural models' robustness, and can also be utilized as data augmentation to increase the models' generalization ability \cite{miyato2016adversarial,Cheng_2019}.\footnote{For more detailed discussion on textual adversarial examples, please refer to recent comprehensive surveys~\cite{zhang2019adversarial,huq2020adversarial,goel2021robustness}.}

\paragraph{White-Box methods} rely on model architecture and parameters being accessible and create adversarial examples directly using a model's gradients.  Unlike image pixel values that are continuous, textual tokens are discrete and cannot be directly modified based on gradients. To this end, adversarial perturbations are added directly to token embeddings or sentence hidden representations \cite{miyato2016adversarial,zhu2019freelb,jiang2019smart,chen-etal-2020-seqvat} which creates ``virtual adversarial examples''. Other approaches vectorize modification operations as the difference of one-hot vectors \cite{ebrahimi2017hotflip,ebrahimi2018adversarial}, or find real word neighbors in a model's hidden representations via its gradients \cite{Cheng_2019}.

\paragraph{Black-Box methods} are usually model-agnostic since they do not require information from a model or its parameters and usually focus on task-specific heuristics for creating adversarial examples. 
For example, by enumerating feasible substitutions on the basis of word similarity and language models, \citet{alzantot2018generating} and \citet{garg-ramakrishnan-2020-bae} select adversarial word replacements which severely influence the predictions from the text classification model. To attack reading comprehension systems, \citet{jia2017adversarial} and \citet{wang-bansal-2018-robust} insert distracting but meaningless sentences at different locations in paragraphs and \citet{ribeiro-etal-2018-semantically} leverage rule-based paraphrasing to produce semantically-equivalent adversarial examples. Likewise, for multi-hop question answering, \citet{jiang-bansal-2019-avoiding} insert shortcut reasoning sentences and \citet{trivedi-etal-2020-multihop} constructed disconnected reasoning example by removing certain supporting facts. For machine translation, \citet{belinkov2017synthetic} attacks character-based models by natural or synthesized typos and  \citet{tan-etal-2020-morphin} further adopt subword morphology level attacks. Similar attacks also help dialogue generation \cite{niu-bansal-2019-automatically} and text summarization \cite{Cheng2020Seq2SickET,Fan2018RobustNA}. Other methods do not rely in editing input text directly; \citet{iyyer2018adversarial} leverage round-trip translation to generate paraphrases in given syntactic templates and  \citet{zhao2017generating} search for adversarial examples in underlying semantic space with GANs \cite{NIPS2014_5423}. Some of these heuristics could be further refined to obtain simple adversarial data augmentation approaches. For example, \citet{mccoy2019right} craft adversarial examples for natural language inference using sophisticated templates which create lexical overlap between the premise and the hypothesis to fool the model. \citet{min2020syntactic}  proposes two simple yet effective adversarial transformations that reverse the position of subject and object or the position of premise and hypothesis.

\subsection{Hidden-Space Augmentation}
\label{subsec:int-DA}
This line of work generates augmented data by manipulating the hidden representations through perturbations such as adding noise or performing interpolations with other data points.
Hidden-space perturbations augment existing data by adding perturbations to the hidden representations of tokens \cite{miyato2016adversarial,zhu2019freelb,jiang2019smart,chen-etal-2020-seqvat,shen2020simple,chen2021hiddencut} or sentences \cite{Hsu_2017,Hsu_2018,Wu2019,malandrakis-etal-2019-controlled}. 

\paragraph{Interpolation-Based Methods.}
Interpolation-based methods create new examples and labels by linear combinations of existing data-label pairs. Given two data-label pairs, 
virtual data-label pairs are created through linear interpolations of the pair of data points. Such interpolation-based methods can generate infinite augmented data in the ``virtual vicinity'' of the original data space, thus improving the generalization performance of models. Interpolation-based methods were first explored in computer vision \cite{zhang2017mixup}, and have more recently been generalized to the text domain \cite{miao2020snippext,chen2020mixtext,cheng-etal-2020-advaug,chen-etal-2020-local} by performing interpolation between original data and token-level augmented data in the output space \cite{miao2020snippext}, between original data and adversarial data in embedding space \cite{cheng-etal-2020-advaug}, or between different training examples in general hidden space \cite{chen2020mixtext}. Different strategies to select samples to mix have also been explored \cite{chen-etal-2020-local, guo-etal-2020-sequence, zhang-etal-2020-seqmix} such as k-nearest-neighbours \cite{chen-etal-2020-local} or sentence composition \cite{guo-etal-2020-sequence}.

\begin{table*}[t]
\centering
\resizebox{2\columnwidth}{!}{%
\begin{tabular}{ccccccc} \toprule
& \multirow{2}{*}{\textbf{Methods}} & \multirow{2}{*}{\textbf{Types}} & \multicolumn{2}{c}{\textbf{News Classification}} & \multicolumn{2}{c}{\textbf{Topic Classification}} \\\cmidrule(lr){4-5}\cmidrule(lr){6-7} 
& &  &  \textbf{AG News} & \textbf{20 Newsgroup} & \textbf{Yahoo Answers} & \textbf{PubMed} \\ \midrule
& None  & - & 78.8(8.9) & 65.2(4.8) & 56.6(9.4) & 63.7(6.1)/49.3(3.9) \\ \cmidrule(lr){2-7}
& SR &   & 79.4(5.9) & 66.1(2.5) & 56.0(10.1) & 62.4(5.7)/48.3(3.9) \\ 
& LM & \multirow{4}{*}{Token}   & 76.8(5.1) & 60.0(14.4) & 56.2(8.4) & 60.9(3.0)/47.4(2.5)  \\ 
\parbox[t]{2mm}{\multirow{4}{*}{\rotatebox[origin=c]{90}{\textbf{Supervised}}}}
 & RI &  & 79.5(4.9) & 66.6(0.6) & 57.3(12.0) & 63.7(4.2)/49.4(2.1) \\ 
& RD &  & 79.6(5.0) & 66.8(3.0) & 58.0(8.3)  & 63.4(5.0)/49.3(1.5)  \\ 
& RS &  & 79.5(5.3) & 64.8(10.8) & 57.1(10.3) & 63.8(7.4)/49.5(3.3) \\ 
& WR &  & 79.7(2.0) & \textbf{67.5(4.2)} & \textbf{59.3(8.9)} & \textbf{64.9(4.9)/49.4(2.5)} \\ \cmidrule(lr){2-7}
& RT & Sentence & \textbf{80.1(4.3)} & 65.1(7.9) & 57.1(9.6) & 60.2(5.1)/46.3(6.4) \\ \cmidrule(lr){2-7}
& ADV &  \multirow{3}{*}{Hidden}  & 78.2 (5.3) & 65.5(1.6) & 53.8(4.89) & 37.4(2.6)/19.9(10.6) \\
& Cutoff & & 79.3(5.0) & 66.6(1.4) & 57.3(9.3) & 60.5(8.3)/46.6(9.4)  \\
& Mixup &  & 80.0 (6.52) & 65.9(3.1) & 57.8(4.19) & 51.4(19.3)/39.8(3.2) \\
\midrule \midrule
& SR & \multirow{5}{*}{Token}   & 69.6(29.3) & 65.7(1.8) & 51.4(9.4) & 59.3(5.9)/43.1(11.9)  \\ 
\parbox[t]{2mm}{\multirow{4}{*}{\rotatebox[origin=c]{90}{\textbf{Semi Supervised}}}} & LM &   & 68.5(13.7) & 68.3(2.1) & 53.2(6.3) & 61.5(6.6)/46.4(4.4)  \\ 
 & RI &  & 65.8(5.5) & 66.7(1.1) & 50.5(3.2) & 61.4(11.3)/44.4(17.4)  \\ 
& RD &  & 73.2(14.0) & 66.1(3.3) & 51.5(7.5) & 59.3(7.1)/46.0(3.8) \\ 
 & RS &  & 71.6(16.6) & 65.0(2.0) & 51.1(7.1) & 64.2(12.1)/46.7(11.5)  \\ 
& WR &  & 74.1(12.3) & \textbf{69.3(2.5)} & 55.6(5.9) & 60.4(7.5)/43.7(14.2) \\ \cmidrule(lr){2-7}
& RT & Sentence & 82.1(8.2) & 68.8(2.4) & 59.8(3.9) & \textbf{64.3(1.2)/49.8(1.9)}\\  \cmidrule(lr){2-7}
& ADV & \multirow{2}{*}{Hidden}  & \textbf{82.3(2.33)} & 66.8(5.9) & 55.9(3.89) & 62.2(10.8)/46.2(9.8) \\ 
& Cutoff &  & 79.9(5.5) & 67.9(0.8) & \textbf{60.1(1.0)} & 62.7(9.0)/48.1(3.2) \\ 
\bottomrule 
\end{tabular}}
\caption{\small Topic Classification and News Classification results with 10 examples. We report the average results across 3 different random seeds with the 95\% confidence interval and \textbf{bold} the best results.. For PubMed, we report the accuracy and F1 score. } \label{Tab:Cls}
\end{table*}

\begin{table*}[t]
\small
\centering
\renewcommand{\tabcolsep}{5pt}
\begin{tabular}{cccccccccc} \toprule
&\multirow{2}{*}{\textbf{Methods}}                                                                        & \multirow{2}{*}{\textbf{Types}}  & \multicolumn{3}{c}{\textbf{Inference}} & \multicolumn{2}{c}{\textbf{Paraphrase}} & \multicolumn{2}{c}{\textbf{Single Sentence}} \\ 
\cmidrule(lr){4-6}\cmidrule(lr){7-8} \cmidrule(lr){9-10}
& &  & \textbf{MNLI} & \textbf{QNLI} & \textbf{RTE}  & \textbf{QQP}  & \textbf{MRPC} & \textbf{SST-2} & \textbf{CoLA}\\ \midrule

&\multirow{1}{*}{None} & -    & 35.2(0.7)	& 51.8(7.0)	& 49.8(3.1)	& 63.9(9.1)	& 61.8(21.2)& 	60.5(13.1)	& 12.9(6.32)  \\  \cmidrule(lr){2-10} 
                                                                                                                                                                
&\multirow{1}{*}{SR}   & \multirow{6}{*}{Token} &35.1(2.3)	&51.4(7.2) &	51.5(3.4)	&61.3(9.7)	&59.7(26.3)	&62.1(17.4)	&7.2(11.6)      \\
                                                                                
&\multirow{1}{*}{LM}     &     &35.3(0.8)	&51.0(8.0)	&49.0(1.4)	&62.4(11)	&61.0(24.3)	&62.8(9.8)	&6.8(15.8)    
                                                                                   \\ \parbox[t]{2mm}{\multirow{5}{*}{\rotatebox[origin=c]{90}{\textbf{Supervised}}}}
&\multirow{1}{*}{RI}     &     &34.9(2.6)	&51.5(8.4)	&\textbf{51.5(1.4)}	&60.6(10.9)	&60.6(25.0)	&63.3(12.2)	&7.8(7.42)      
                                                                                 \\ 
&\multirow{1}{*}{RD}       &    &\textbf{35.5(2.1)}	&51.1(8.4)	&50.9(2.4)	&62.4(11.3)	&61.2(22.0)	&59.7(18.4)	&7.1(16.6)      
                                                                                  \\ 
&\multirow{1}{*}{RS}       &      &35.1(1.1)	 &51.5(7.0)	 &50.9(5.0)	 &\textbf{62.6(6.7)}	 &\textbf{63.2(22.5)}	 &61.2(10.8)	 &5.2(17.0)    
                                                                                \\ 
&\multirow{1}{*}{WR}     &    &34.5(2.6)	&\textbf{52.0(3.8)}	&50.0(0.9)	&60.6(10.2)	&61.0(25.3)	&61.8(12.5)	&7.0(10.6)       
                                                                                  \\ \cmidrule(lr){2-10}

&\multirow{1}{*}{RT} & Sentence &   35.3(0.5)	&51.1(9.6)	&50.8(4.4)	&60.5(17.8)	&61.8(23.7)	&62.0(1.99)	&8.37(8.35)    
                                                                                   \\ \cmidrule(lr){2-10}
                                                                                   
&\multirow{1}{*}{ADV} & \multirow{3}{*}{Hidden}    &33.3(4.7)	&49.7(1.8)	&48.3(12.1)	&57.5(24.7)	&61.5(21.5)	&53.3(13.07)	&1.37(4.66)     
                                                                                   \\

&\multirow{1}{*}{Cutoff} &     &35.1(2.3)	&51.4(8.3)	&52.2(3.6)	&62.6(8.8)	&61.0(21.2)	&\textbf{63.5(8.45)}	&\textbf{12.4(9.58)}   \\  

&\multirow{1}{*}{Mixup} &        &32.6(3.5)	 &49.9(1.4)	 &49.8(9.2)	 &63.0(0.3)	 &62.1(19.8)	 &62.3(12.3)	 &4.03(8.68)  
                                                                                    
                                                                                 \\ \midrule \midrule

&\multirow{1}{*}{SR}   & \multirow{6}{*}{Token}    & 35.6(1.0)	&52.1(4.5)	&\textbf{52.9(5.4)}	&53.5(10.7)	&68.1(4.0)	&61.8(37.9)	&6.65(5.69)     \\ \parbox[t]{2mm}{\multirow{4}{*}{\rotatebox[origin=c]{90}{\textbf{Semi-Supervised}}}}
                                                                                
&\multirow{1}{*}{LM}     &    &35.0(3.3)	&52.5(4.2)	&50.2(6.5)	&47.9(34.1)	&68.4(3.8)	&57.3(14.2)	&6.38(6.3)
                                                                                   \\ 
&\multirow{1}{*}{RI}     &    &\textbf{35.8(1.7)}	&52.1(4.1)	&50.7(1.4)	&59.6(5.1)	&64.9(8.9)	&58.3(14.8)	&6.55(0.91)
                                                                                 \\ 
&\multirow{1}{*}{RD}       &   &35.2(0.5)	&52.1(5.2)	&52.6(4.9)	&56.1(16.0)	&62.4(30.6)	&55.7(16.4)	&4.33(10.9)
                                                                                  \\ 
&\multirow{1}{*}{RS}       &     &34.6(2.5)	&52.1(6.2)	&51.5(3.7)	&49.8(7.9)	&63.2(22.5)	&55.2(15.3)	&7.77(11.77)   
                                                                                \\ 
&\multirow{1}{*}{WR}     &    &34.8(2.5)	&52.1(4.1)	&50.9(1.8)	&51.8(16.0)	&63.1(23.5)	&54.8(13.8)	&5.43(17.8)
                                                                                  \\ \cmidrule(lr){2-10}

&\multirow{1}{*}{RT} & Sentence  &35.3(2.7)	&\textbf{52.7(4.8)}	&51.6(4.1)	&\textbf{63.9(7.5)}	&62.2(12.5)	&\textbf{61.9(20.8)}	&\textbf{11.6(14.5)}
                                                                                   \\ \cmidrule(lr){2-10}
                                                                                   
&\multirow{1}{*}{ADV} & \multirow{2}{*}{Hidden}   &36.2(8.9)	&50.6(1.9)	&50.9(6.8)	&59.1(14.7)	&63.9(9.1)	&53.1(5.0)	&7.64(25.1)     
                                                                                   \\

&\multirow{1}{*}{Cutoff} &   &35.3(2.8)	&52.5(4.3)	&51.7(6.5)	&62.9(9.9)	&\textbf{68.6(4.4)}	&54.3(9.8)	&4.11(11.8)
                                                                                 \\ \bottomrule 

\end{tabular}
\caption{\small GLUE results with 10 labeled examples per class. We report the average results across 3 different random seeds with the 95\% confidence interval and \textbf{bold} the best results.} \label{Tab:Glue_10}
\end{table*}

We summarize the preceding overview of recent widely-used data augmentation methods in Table \ref{da-sum}, characterizing them with respect to augmentation levels, the diversity of generated data, and their applicable tasks.

\section{Consistency Training with DA}
While data augmentation (DA) can be applied in the supervised setting to produce better results when only a small labeled training dataset is available, data augmentation is also commonly used in semi-supervised learning (SSL).
SSL is an alternative approach for learning from limited data that provides a framework for taking advantage of unlabeled data.
Specifically, SSL assumes that our training set comprises labeled examples in addition to unlabeled examples drawn from the same distribution.
Currently, one of the most common methods for performing SSL with deep neural networks is ``consistency regularization'' \cite{bachman2014learning,tarvainen2017mean}.
Consistency regularization-based SSL (or ``consistency training'' for short) regularizes a model by enforcing that its output doesn't change significantly when the input is perturbed.
In practice, the input is perturbed by applying data augmentation, and consistency is enforced through a loss term that measures the difference between the model's predictions on a clean input and a corresponding perturbed version of the same input.

Formally, let $f_\theta$ be a model with parameters $\theta$, $f_{\hat{\theta}}$ be a fixed copy of the model where no gradients are allowed to flow, $x_l$ be a labeled datapoint with label $y$, $x_u$ be an unlabeled datapoint, and $\alpha(x)$ be a data augmentation method.
Then, a typical loss function for consistency training is 
\begin{align*}
\mathrm{CE}(f_\theta(x_l), y) + \lambda_u \mathrm{CE}(f_{\hat{\theta}}(x_u), f_{\theta}(\alpha(x_u)))  
\end{align*}
where $\mathrm{CE}$ is the cross entropy loss and $\lambda_u$ is a tunable hyperparameter that determines the weight of the consistency regularization term.
In practice, various other measures have been used to minimize the difference between $f_{\hat{\theta}}(x_u)$ and $f_{\theta}(\alpha(x_u))$, such as the KL divergence \cite{miyato2018virtual,xie2019unsupervised} and the mean-squared error \cite{tarvainen2017mean,laine2016temporal,berthelot2019mixmatch}. 
Because gradients are not allowed to flow through the model when it was fed the clean unlabeled input $x_u$, this objective can be viewed as using the clean unlabeled datapoint to generate a synthetic target distribution for the augmented unlabeled datapoint.

\citet{xie2019unsupervised} showed that consistency training can be effectively applied to semi-supervised learning for NLP.
To achieve stronger results, they introduce several other tricks including confidence thresholding, training signal annealing, and entropy minimization.
Confidence thresholding applies the unsupervised loss only when the model assigns a class probability above a pre-defined threshold.
Training signal annealing prevents the model from overfitting on easy examples by applying the supervised loss only when the model is less confident about predictions.
Entropy minimization trains the model to output low-entropy (highly-confident) predictions when fed unlabeled data.
We refer the reader to \cite{xie2019unsupervised} for more details on these tricks.

\section{Empirical Experiments}
\subsection{Datasets and Experiment Setup}
To provide a quantitative comparison of the DA methods we have surveyed, we experiment with 10 of the most commonly used and model-agnostic augmentation techniques from different levels in Table \ref{da-sum}, including: (i) \textit{Token-level augmentation}: \textbf{S}ynonym  \textbf{R}eplacement (\textbf{SR}) \cite{kolomiyets-etal-2011-model,Yang2015That}, Word Replacement based on \textbf{L}anguage \textbf{M}odel (\textbf{LM} \cite{kumar2020data}, \textbf{R}andom \textbf{I}nsertion (\textbf{RI}) \cite{wei2019eda,miao2020snippext}, \textbf{R}andom \textbf{Deletion} (\textbf{RD}) \cite{wei2019eda}, \textbf{R}andom \textbf{S}wapping (\textbf{RS}) \cite{wei2019eda}, and \textbf{W}ord \textbf{R}eplacement (\textbf{WR}) based on TF-IDF in \textit{Vocabulary Set} \cite{xie2019unsupervised}; (ii) \textit{Sentence-level augmentation}: \textbf{R}oundtrip \textbf{T}ranslation (\textbf{RT}) \cite{xie2019unsupervised,chen2020mixtext}; (iii) \textit{Hidden-space Augmentation}: \textbf{Adv}ersarial training (\textbf{ADV}) \cite{goodfellow2014explaining}, \textbf{Cutoff} \cite{shen2020simple}, and \textbf{Mixup} in the embedding space \cite{zhang2017mixup}. Most aforementioned techniques are not label-dependent (except mixup), thus can be applied directly to unlabeled data. 
    
We test them on different types of benchmark datasets including: (i) news classification tasks including AG News \cite{ZhangZL15} and 20 Newsgroup \cite{10.5555/645526.657278}; (ii) topic classification tasks including Yahoo Answers \cite{Chang:2008:ISR:1620163.1620201} and PubMed
news classification (\cite{ZhangZL15} (iii) inference tasks including MNLI, QNLI and RTE \cite{wang-etal-2018-glue}; (iv) similarity and paraphrase tasks including QQP and MRPC  \cite{wang-etal-2018-glue}; and (v) single-sentence tasks including SST-2 and CoLA \cite{wang-etal-2018-glue}.

For all datasets, we experiment with 10 labeled data points per class \footnote{The results for 100 labeled data points per class are shown in the Appendix.} in a supervised setup, and an additional 5000 unlabeled data points per class in the semi-supervised setup. We use \textit{BERT\textsubscript{base}} \cite{devlin2018bert} as the base language model and use the same hyper-parameters across all datasets/methods.  We utilize accuracy as the evaluation metric for all datasets except for CoLA (which uses Matthews correlation) and PubMed (which uses accuracy and Macro-F1 score). Because the performance can be heavily dependent on the specific datapoints chosen \cite{sohn2020fixmatch}, for each dataset, we sample labeled data from the original dataset with 3 different seeds to form different training sets, and report the average result. For every setup, we fine-tune the model with the same seed as the dataset seed (in contrast to many works which report the max across different seeds).  The detailed experimental setup is described in the Appendix.

\subsection{Results}
\paragraph{News/Topic Classification Tasks.} The results 
are shown in Table~\ref{Tab:Cls}. We observe that in supervised settings, \textit{token-level augmentations} work the best. Specifically, word replacement works well, getting the highest or second highest score every time; in the semi-supervised settings, \textit{sentence level augmentations} (round-trip translation) works the best, getting the highest or second highest score every time. This makes sense since for many classification tasks, multiple words indicate the label, and so dropping several words will not affect the label. 

\paragraph{Inference Tasks.} As shown in Table~\ref{Tab:Glue_10}, we observe that \textit{token-level augmentations} work the best overall (e.g., random insertion, random deletion, and word replacement) for both supervised and semi-supervised settings. This is a bit surprising since the inference tasks usually heavily depend on several words, and changing these words can easily change the label for inferene tasks.

\paragraph{Similarity and Paraphrase Tasks.} From Table~\ref{Tab:Glue_10}, in the supervised settings, we observe that  \textit{token-level augmentations} (random swapping) achieve the best performances, while \textit{hidden space augmentations} work well in semi-supervised settings, with cutoff performing the best on average. This makes sense since for paraphrasing tasks, augmenting the text usually consists of paraphrases, and so can easily change whether two texts are paraphrases of each other.

\paragraph{Single Sentence Tasks.} Based on the single-sentence tasks results in Table~\ref{Tab:Glue_10}, \textit{hidden space augmentations} (cutoff) provides the biggest boost in performance in supervised settings, while in semi-supervised settings, \textit{sentence level augmentations} (roundtrip translation) works best. We note most augmentation methods hurt performance on CoLA, a task for judging grammatical acceptability. This could be caused by the fact that most of augmentation methods try to preserve meaning and not grammatical correctness.

Overall, \textbf{no single augmentation works the best for every task in the supervised or semi-supervised setting}. However, several overall conclusions can be made: first, augmentation does not always improve performance, and can sometimes hurt performances, even in the semi-supervised setting. This suggests that we may need to design different augmentations for different tasks. Second, token-level augmentations (especially word replacement and random swapping) work well in general for supervised learning, especially when there is extremely limited labeled data. Third, round-trip translation usually works the best for semi-supervised learning, showing the most consistent gains. However, if the computation is limited, cutoff may be a better choice. 

\section{Other Limited Data Learning Methods}
\label{other-methods}

This work mainly focuses on data augmentation and semi-supervised learning (consistency regularization) in NLP; however, there are other orthogonal directions for tackling the problem of learning with limited data.
For completeness, we summarize this related work below.

\paragraph{Low-Resourced Languages.}
Most languages lack large monolingual or parallel corpora, or sufficient manually-crafted linguistic resources for building statistical NLP applications \cite{garrette-baldridge-2013-learning}. Researchers have therefore developed a variety of methods for improving performance on low-resource languages, including cross-lingual transfer learning which transfers models from resource-rich to resource-poor languages \cite{do-gaspers-2019-cross,lee2019crosslingual,schuster-etal-2019-cross-lingual}, few/zero-shot learning \cite{johnson-etal-2017-googles,blissett-ji-2019-zero,pham-etal-2019-improving,Abad_2020} which uses only a few examples from the low-resource domain to adapt models trained in another domain, and polyglot learning \cite{cotterell-heigold-2017-cross,tsvetkov-etal-2016-polyglot,mulcaire-etal-2019-low,lample2019cross} which combines resource-rich and resource-poor learning using an universal language representation. 
\paragraph{Other Methods for Semi-Supervised Learning.}
Semi-supervised learning methods further reduce the dependency on labeled data and enhance the models when there is only limited labeled data available. These methods use large amounts of unlabeled data in the training process, as unlabeled data is usually cheap and easy to obtain compared to labeled data. In this paper, we focus on consistency regularization, while there are also other widely-used methods for NLP including self-training \cite{yarowsky1995unsupervised,zhang-zong-2016-exploiting,he2019revisiting,lin2020triggerner}, generative methods \cite{xu2016variational,10.5555/3305890.3306082,kingma2014semisupervised,cheng2016semisupervised}, and co-training \cite{blum1998combining,clark2018semisupervised,cai-lapata-2019-semi}.

\paragraph{Few-shot Learning.}
Few-shot learning is a broad technique for dealing with tasks with less labeled data based on prior knowledge. Compared to semi-supervised learning which utilizes unlabeled data as additional information, few-shot learning leverages various kinds of prior knowledge such as pre-trained models or supervised data from other domains and modalities \cite{Wang_2020}. While most work on few-shot focuses on computer vision, few-shot learning has recently seen increasing adoption in NLP \cite{han-etal-2018-fewrel,rios-kavuluru-2018-shot,hu-etal-2018-shot,herbelot-baroni-2017-high}. To better leverage pre-trained models, PET \cite{schick2020pet, schick202ipet} converts the text and label in an example into a fluent sentence, and then uses the probability of generating the label text as the class logit, outperforming GPT3 for few shot learning \cite{GPT3}.  How to better model and incorporate prior knowledge to handle few-shot learning for NLP remains an open challenge and has the potential to significantly improve model performance with less labeled data.

\section{Discussion and Future Directions} 
\label{future_directions}
In this work, we empirically surveyed data augmentation methods for limited-data learning in NLP and compared them on 11 different NLP tasks. Despite the success, there are still certain challenges that need to be tackled for improve their performance. This section highlights some of these challenges and discusses future research directions.

\paragraph{Theoretical Guarantees and Data Distribution Shift.}
Current data augmentation methods for text typically assume that they are label-preserving and will not change the data distribution. However, these assumptions are often not true in practice, which can result in noisy labels or a shift in the data distribution and consequently a decrease in performance or generalization (e.g., QQP in Table~\ref{Tab:Glue_10}). Thus, providing theoretical guarantees that augmentations are label- and distribution-preserving under certain conditions would ensure the quality of augmented data and further accelerate the progress of this field. 

\paragraph{Automatic Data Augmentation.}
Despite being effective, current data augmentation methods are generally manually-designed. Methods for automatically selecting the appropriate types of data augmentation still remain under-investigated. Although certain augmentation techniques have been shown effective for a particular task or dataset, they often do not transfer well to other datasets or tasks \cite{cubuk2019autoaugment}, as shown in Table~\ref{Tab:Glue_10}. For example, paraphrasing works well for general text classification tasks, but may fail for some subtle scenarios like classifying bias because paraphrasing might change the label in this setting. 
Automatically learning data augmentation strategies or searching for an optimal augmentation policy for given datasets/tasks/models could enhance the generalizability of data augmentation techniques \cite{maharana2020dataaug}.

\section*{Acknowledgments}
We would like to thank the members of Georgia Tech SALT and UNC-NLP groups for their feedback. This work is supported by grants from Amazon and Salesforce, ONR Grant N00014-18-1-2871, DARPA YFA17-D17AP00022.

\bibliographystyle{acl}
\bibliography{acl2021}

\appendix

\section{Experimental Setup}

We train our models on NVIDIA 2080ti and NVIDIA V-100 gpus. Supervised experiments take 20 minutes, and semi-supervised experiments take two hours. The BERT-base model has 100M parameters. We use the same hyperaparameter across all datasets, and so only use the validation set to find the best model checkpoint. We use a learning rate of $2e^{-5}$, batch size of $16$, ratio of unlabeled to labeled data of 3, and dropout ratio of $0.1$ for different augmentation methods.

\section{Results for 100 Labeled Data per Class}

\paragraph{News/Topic Classification Tasks} The results 
are shown in Table~\ref{Tab:tc_100}. We observe that overall, in both the supervised settings and semi-supervised setting, all the methods perofrmly similarly, with 2 points of each other. This indicates that data augmentation methods work well with limited labeled data, and with more labeled data, its effectiveness is removed. 

\paragraph{Inference Tasks} As shown in Table~\ref{Tab:Glue_100}, we observe that most augmentation methods hurt the performance in both the supervised and semi-supervised setting, with a greater drop in performance in the semi-supervised setting.

\paragraph{Similarity and Paraphrase Tasks} Similar to \textit{inference tasks}, we observe in Table~\ref{Tab:Glue_100} that most augmentation methods hurt the performance in both the supervised and semi-supervised setting, with a greater drop in performance in the semi-supervised setting. 

\paragraph{Single Sentence Tasks} Unlike \textit{inference tasks} and \textit{paraphrase tasks}, augmentations methods help performance, as seen in Table~\ref{Tab:Glue_100}, except for CoLA. We hypothesize the reason is because most augmentatiom methods seek to preserves meaning, not grammatical correctness, which is what CoLA measures. In the supervised and semi-supervised setting, hidden level augmentations work well, with cutoff performing the best.

\begin{table*}[t]
\centering
\resizebox{2\columnwidth}{!}{%
\begin{tabular}{ccccccc} \toprule
& \multirow{2}{*}{\textbf{Methods}} & \textbf{Types} & \multicolumn{2}{c}{\textbf{News Classification}} & \multicolumn{2}{c}{\textbf{Topic Classification}} \\ \cmidrule(lr){4-5}\cmidrule(lr){6-7} 
& & - &  \textbf{AG News} & \textbf{20 Newsgroup} & \textbf{Yahoo Answers} & \textbf{PubMed} \\ \midrule 
& None  & - & 87.9(1.05) & 79.5(0.3) & 68.6(0.71) & 75.2(1.5)/59.5(2.0)  \\ \cmidrule(lr){2-7}
& SR &  & 88.5(0.87) & 80.0(2.2) & \textbf{69.7(1.62)} & 76.5(1.0)/60.7(0.7) \\ 
& LM & \multirow{4}{*}{Token}  & 88.1(1.00) & \textbf{80.5(1.8)} & 68.8(3.2) & 75.8(2.5)/59.9(1.7) \\ 
\parbox[t]{2mm}{\multirow{4}{*}{\rotatebox[origin=c]{90}{\textbf{Supervised}}}}
 & RI &  & 88.0(2.08) & 80.1(3.1) & 69.1(1.68) & 76.2(2.9)/60.3(1.7) \\ 
& RD &  & 88.1(0.84) & 80.2(2.9) & 68.7(2.2) & \textbf{76.9(0.6)/60.9(0.6)}  \\ 
& RS &  & 88.4(0.97) & 79.5(2.1) & 69.0(2.03) & 76.6(0.2)/60.6(0.7) \\ 
& WR &  & 87.9(1.19) & 79.3(2.5) & 69.4(5.89) & 76.4(1.8)/60.4(1.6)\\ \cmidrule(lr){2-7}
& RT & Sentence & 88.3(0.17) & 80.4(0.7) & 68.8(1.88) & 76.1(0.5)/60.3(0.5) \\ \cmidrule(lr){2-7}
& ADV &  \multirow{3}{*}{Hidden} & 87.6(0.33) & 78.5(1.4) & 67.4(0.74) & 75.6(4.0)/59.8(3.5) \\
& Cutoff & & 88.3(0.38) & 79.8(1.0) & 68.7(0.47) & 75.9(1.3)/60.1(0.7)  \\
& Mixup &  & \textbf{88.6(1.31)} & {80.5(3.4)} & 68.27(1.76) & 74.8(1.8)/59.2(0.2) \\
\midrule \midrule
& SR & \multirow{5}{*}{Token} & 88.8(0.95) & 81.2(8.4) & 68.8(1.3) & 76.6(1.5)/60.7(1.8) \\ 
\parbox[t]{2mm}{\multirow{4}{*}{\rotatebox[origin=c]{90}{\textbf{Semi Supervised}}}} & LM &   & 88.4(1.87) & 81.4(1.0) & 68.8(1.8) & 76.4(1.3)/60.4(0.7) \\ 
 & RI &  & 88.4(1.45) & 80.3(3.0) & 68.4(2.64) & 76.8(1.2)/60.7(1.1)  \\ 
& RD &  & 88.7(0.5) & 80.5(0.8) & 68.8(1.66) & \textbf{77.1(1.0)/61.2(1.5)}\\ 
 & RS &  & 88.5(1.35) & 80.9(2.2) & 68.7(1.67) & 76.9(1.7)/61.0(1.5) \\ 
& WR &  & 87.7(1.35) & 81.5(1.3) & 68.7(1.2) & 76.5(0.5)/60.6(1.0)\\ \cline{2-7}
& RT & Sentence & 88.7(0.40) & \textbf{81.7(1.0)} & \textbf{69.7(1.06)} & 77.0(1.2)/61.6(1.1) \\  \cmidrule(lr){2-7}
& ADV & \multirow{2}{*}{Hidden} & 88.0(1.04) & 80.4(2.9) & 68.9(1.74) & 76.7(1.5)/60.9(1.2)\\ 
& Cutoff &  & \textbf{88.9(0.25)} & 81.3(4.6) & 69.3(1.76) & 76.7(2.1)/60.7(3.1) \\ 
\bottomrule 
\end{tabular}}
\caption{\small Topic Classification and News Classification results with 100 examples. We report the average results across 3 different random seeds with the 95\% confidence interval and \textbf{bold} the best results.. For PubMed, we report the accuracy and F1 score. } \label{Tab:tc_100}
\end{table*} 
\begin{table*}[t]
\small
\centering
\renewcommand{\tabcolsep}{5pt}
\begin{tabular}{cccccccccc} \toprule
&\multirow{2}{*}{\textbf{Methods}}                                                                        & \multirow{2}{*}{\textbf{Types}}  & \multicolumn{3}{c}{\textbf{Inference}} & \multicolumn{2}{c}{\textbf{Paraphrase}} & \multicolumn{2}{c}{\textbf{Single Sentence}} \\ \cmidrule(lr){4-6}\cmidrule(lr){7-8} \cmidrule(lr){9-10}
& &  & \textbf{MNLI} & \textbf{QNLI} & \textbf{RTE}  & \textbf{QQP}  & \textbf{MRPC} & \textbf{SST-2} & \textbf{CoLA}\\ \midrule

&\multirow{1}{*}{None} & -    &45.0(6.9)	&63.2(10.7)	&59.9(3.1)	&71.0(2.6)	&68.1(7.4)	&82.7(4.0)	&28.7(9.5)  \\\cmidrule(lr){2-10}
                                                                                                                                                                
&\multirow{1}{*}{SR}   & \multirow{6}{*}{Token} &44.6(7.2)	&62.9(9.4)	&61.0(10.0)	&68.9(2.2)	&66.7(4.4)	&84.0(1.9)	&24.6(5.1)     \\
                                                                                
&\multirow{1}{*}{LM}     &     &45.4(6.2)	&60.6(7.7)	&\textbf{61.5(9.1)}	&69.6(1.7)	&67.2(2.8)	&83.8(3.1)	&18.5(9.7)
                                                                                   \\ \parbox[t]{2mm}{\multirow{5}{*}{\rotatebox[origin=c]{90}{\textbf{Supervised}}}}
&\multirow{1}{*}{RI}     &     &\textbf{45.8(7.5)}	&\textbf{64.2(10.7)}	&60.0(11.3)	&69.2(0.6)	&69.1(4.8)	&84.3(1.4)	&27.3(19.9)
                                                                                 \\ 
&\multirow{1}{*}{RD}       &    &43.7(8.4)	&63.6(9.4)	&59.2(9.0)	&69.2(1.5)	&69.2(5.5)	&82.3(2.05)	&20.2(21.5)
                                                                                  \\ 
&\multirow{1}{*}{RS}       &     &42.4(6.2)	&63.3(9.1)	&57.8(11.9)	&68.3(1.6)	&69.0(3.4)	&82.5(5.0)	&24.3(20.8)    
                                                                                \\ 
&\multirow{1}{*}{WR}     &    &44.6(6.3)	&61.6(8.8)	&57.8(9.3)	&66.7(1.8)	&66.9(6.4)	&83.5(1.9)	&17.7(23.3)
                                                                                  \\\cmidrule(lr){2-10}

&\multirow{1}{*}{RT} & Sentence &44.8(7.8)	&59.0(7.6)	&60.4(5.7)	&\textbf{69.9(4.0)}	&\textbf{69.6(1.6)}	&84.3(3.27)	&19.2(7.63)
                                                                                   \\ \cmidrule(lr){2-10}
                                                                                   
&\multirow{1}{*}{ADV} & \multirow{3}{*}{Hidden}    &39.1(10.9)	&50.1(3.1)	&57.3(8.7)	&63.7(1.9)	&68.7 (6.3)	&69.8(5.3)	&16.5(9.2)
                                                                                   \\

&\multirow{1}{*}{Cutoff} &     &44.9(5.5)	&63.0(10.2)	&59.3(8.8)	&69.9(0.7)	&66.5(1.3)	&\textbf{84.7(0.9)}	&\textbf{26.0(16.3)}   \\  

&\multirow{1}{*}{Mixup} &        &35.7(7.3)	&51.4(4.4)	&60.5(6.52)	&64.5(5.4)	&67.9 (7.1)	&83.5(3.4)	&20.1(18.8)
                                                                                    
                                                                                 \\ \midrule  \midrule

&\multirow{1}{*}{SR}   & \multirow{6}{*}{Token}    &42.9(7.3)	&60.1(6.2)	&58.5(9.7)	&65.0(6.0)	&67.6(3.1)	&85.1(3.5)	&18.9(6.7)  \\ \parbox[t]{2mm}{\multirow{4}{*}{\rotatebox[origin=c]{90}{\textbf{Semi-Supervised}}}}
                                                                                
&\multirow{1}{*}{LM}     &   &43.7(4.5)	&60.9(10.4)	&56.9(8.3)	&59.3(12.0)	&70.0(4.4)	&83.9(4.1)	&21.7(6.8)
                                                                                   \\ 
&\multirow{1}{*}{RI}     &    &44.7(4.6)	&62.5(10.5)	&56.0(6.3)	&68.3(0.1)	&67.0(3.9)	&84.2(3.0)	&23.0(10.3)
                                                                                 \\ 
&\multirow{1}{*}{RD}       &   &41.4(2.9)	&59.4(6.4)	&56(0.0)	&\textbf{69.3(2.8)}	&70.4(7.4)	&83.6(2.3)	&13.1(6.1)
                                                                                  \\ 
&\multirow{1}{*}{RS}       &     &40.3(2.0)	&60.3(8.7)	&56.4(11.6)	&66.8(2.3)	&69.0(3.4)	&84.5(3.6)	&19.4(2.7)
                                                                                \\ 
&\multirow{1}{*}{WR}     &   &43.9(3.1)	&60.5(8.8)	&56.3(7.1)	&65.4(4.3)	&67.2(2.1)	&83.3(4.5)	&16.9(6.2)
                                                                                  \\ \cmidrule(lr){2-10}

&\multirow{1}{*}{RT} & Sentence   & \textbf{45.4(7.7)}	 & \textbf{63.8(5.0)}	 & \textbf{59.9(9.1)}	 & 68.3(2.9)	 & 67.5(0.7)	 & \textbf{83.9(1.7)}	 & 20.4(3.6)                         \\ \cmidrule(lr){2-10}
                                                                                   
&\multirow{1}{*}{ADV} & \multirow{2}{*}{Hidden}   &44.1(3.4)	&58.1(4.0)	&58.6(5.2)	&63.0(10.8)	&67.6(5.2)	&80.0(7.3)	&13.5(7.8)
                                                                                   \\

&\multirow{1}{*}{Cutoff} &    &42.7(4.2)	 &60.3(7.4)	 &57.9(12.6)	 &67.2(4.4)	 &\textbf{71.4(2.0)}	 &82.5(5.4)	 &\textbf{23.9(2.7)}
                                                                                 \\ \bottomrule 

\end{tabular}
\caption{\small GLUE results with 100 labeled examples per class. We report the average results across 3 different random seeds with the 95\% confidence interval and \textbf{bold} the best results.} \label{Tab:Glue_100}
\end{table*} 

\section{Case Study}

We analyze several data augmentation methods and check whether the label is preserved for these and if this affects its performance. We look at 25 examples for the best performing data augmentation method and the worst performing data augmentation method for \textit{20 News Group} and \textit{RTE}. For \textit{20 News Group}, \textit{Random Deletion} was the best performing, and \textit{Language Model} was the worst performing. In both cases, there were no examples where the label flipped, which makes sense since the input is usually several paragraphs with multiple references to the topic. Several examples are shown in Appendix. For \textit{RTE}, \textit{Language Model} was the worst performing and \textit{Cutoff} was the best performing augmentation. \textit{Language Model} flipped $24\%$ of the labels with $4\%$ uncertain, while \textit{Cutoff} flipped $4\%$ of the labels with $12\%$ uncertain. We show several examples of when the label flipped for RTE in the Table~\ref{Tab:rte_ex}.

\begin{table*}
\begin{tabular}{p{0.3\linewidth} p{0.3\linewidth} p{0.3\linewidth} } \toprule
\textbf{Original}  & \textbf{Cutoff (Best)} & \textbf{Language Model (Worst)} \\ 
\midrule
\midrule
Sentence 1: The Walt Disney Co. donated one of the world's most significant private collections of African artwork, yesterday, to the Smithsonian's National Museum of African Art. & Sentence 1: The Walt Disney Co. donated one of the world's most significant private collections of African artwork, yesterday, to the Smithsonian's National Museum of African one & Sentence 1: The Walt Disney Co. donated one of the world's most significant private collections of African artwork [PAD] [PAD] [PAD] to the Smithsonian's National Museum of African Art. \\
Sentence 2: Disney gave the Smithsonian a trove of sought-after African art. & Sentence 2: Disney gave the Smithsonian a trove of south African art. & Sentence 2: Disney gave the Smithsonian a trove of [PAD] African art. \\
\midrule
Entailment & Entailment & Not Entailment \\
\midrule
\midrule

Sentence 1: An explosion, followed by a raging fire, demolished a plastics factory, killing at least three people and injuring at least 37. & Sentence 1: An explosion, followed by a raging fire, demolished a the factory, killing at least three people and injuring at least 37. & Sentence 1: An explosion, followed by [PAD] [PAD] fire, demolished a plastics factory, killing at least three people and injuring at least 37. \\
Sentence 2: A massive blast at a plastics factory killed at least two people. & Sentence 2: A massive blast at a plastics factory killed at shot two people. & Sentence 2: A massive blast at a plastics [PAD] killed at least two people. \\
\midrule 
Entailment & Entailment & Not Entailment \\
\midrule
\midrule

Sentence 1: The prize is named after Alfred Nobel, a pacifist and entrepreneur who invented dynamite in 1866. Nobel left much of his wealth to establish the award, which has honoured achievements in physics, chemistry, medicine, literature and efforts to promote peace since 1901. & Setence 1: The prize is named after Alfred Nobel, a pacifist and entrepreneur who invented dynamite in 1866. Nobel left much of his wealth to establish the nobel which has honoured achievements in physics, chemistry, medicine, literature and efforts to promote peace since 1901. & The prize is named after Alfred Nobel, a pacifist and entrepreneur who invented dynamite in 1866 . Nobel left much of his wealth [PAD] [PAD] [PAD] [PAD], which has honoured achievements in physics, chemistry, medicine, literature and efforts to promote peace since 1901. \\
\midrule 
Sentence 2: Alfred Nobel invented dynamite in 1866. & Sentence 2: Alfred Nobel invented dynamite in 1866. & Sentence 2: Alfred Nobel invented dynamite in 1866. \\
\midrule
Entailment & Entailment & Not Entailment \\ \bottomrule

\end{tabular}
\caption{Examples of different data augmentation methods on RTE and whether they preserve the original label or not}
\label{Tab:rte_ex}
\end{table*}

\end{document}